\documentclass[runningheads]{llncs}

\usepackage{graphicx}
\usepackage{hyperref}
\usepackage{xcolor}
\usepackage{amsmath}
\usepackage{amssymb}
\usepackage{booktabs}
\usepackage{microtype}

\input glyphtounicode
\title{Projective Latent Interventions for Understanding and Fine-tuning Classifiers}
\titlerunning{Projective Latent Interventions}
\author{%***%
    Andreas Hinterreiter\inst{1,2}  \and %\orcidID{0000-0003-4101-5180}
    Marc Streit\inst{2}             \and %\orcidID{0000-0001-9186-2092}
    Bernhard Kainz\inst{1}               %\orcidID{0000-0002-7813-5023}
}
\authorrunning{A. Hinterreiter et al.}

\institute{%
Biomedical Image Analysis Group,
    Imperial College,
    UK\\
    \email{\{a.hinterreiter, b.kainz\}@imperial.ac.uk} \and
Istitute of Computer Graphics,
    Johannes Kepler University Linz,
    Austria\\
    \email{\{andreas.hinterreiter, marc.streit\}@jku.at}
}

\newcommand{\method}{Projective Latent Interventions}
\newcommand{\shortmethod}{PLIs}

\begin{document}

\maketitle

%\vspace*{-2ex}

\begin{abstract}
High-dimensional latent representations learned by neural network classifiers are notoriously hard to interpret.
Especially in medical applications, model developers and domain experts desire a better understanding of how these latent representations relate to the resulting classification performance.
We present \method{} (\shortmethod), a technique for retraining classifiers by back-propagating manual changes made to low-dimensional embeddings of the latent space.
The back-propagation is based on parametric approximations of  \(t\)-distributed stochastic neighbourhood embeddings.
\shortmethod{} allow domain experts to control the latent decision space in an intuitive way in order to better match their expectations.
For instance, the performance for specific pairs of classes can be enhanced by manually separating the class clusters in the embedding.
We evaluate our technique on a real-world scenario in fetal ultrasound imaging.

\keywords{Latent space \and Non-linear embedding \and Image classification.}
\end{abstract}

\section{Introduction}

The interpretation of classification models is often difficult due to a high number of parameters and high-dimensional latent spaces.
Dimensionality reduction techniques are commonly used to visualise and explain latent representations via low-dimensional embeddings.
These embeddings are useful to identify problematic classes, to visualise the impact of architectural changes, and to compare new approaches to previous work.
However, there is a lot of debate about how well such mappings represent the actual decision boundaries and the resulting model performance.

\begin{figure}[tb]
 \centering
 \includegraphics[width=\textwidth]{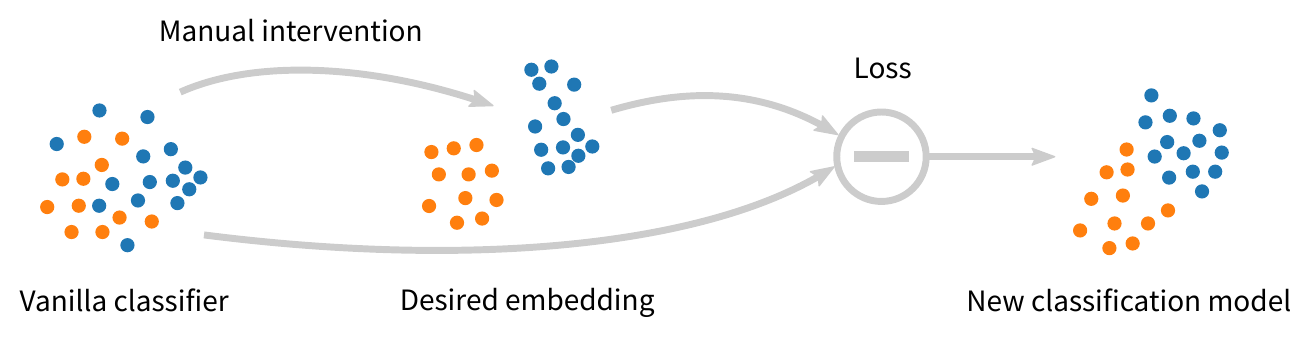}
 \caption{\shortmethod{} define a desired embedding, which is subsequently used to inform the training or fine-tuning process of a classification model in an end-to-end way.}
 \label{fig:method_outline}
\end{figure}

In this work, we aim to change the paradigm of passive observation of mappings to active interventions during the training process.
We argue that such interventions can be useful to mentally connect the embedded latent space with the classification properties of a classifier.
We show that in some situations, such as class-imbalanced problems, the manual interventions can also be used for fine-tuning and targeted performance gains.
This means that practitioners can prioritise the decision boundary for certain classes over the others simply by manipulating the embedded latent space.
The overall idea of our work is outlined in Fig.~\ref{fig:method_outline}.
% In contrast, other ways of prioritising classes, such as weighting the loss, often lead to labour-intensive hyper-parameter tuning~\cite{johnson2019survey}.
We use a neural-network-based parametric implementation of \(t\)-distributed stochastic neighbourhood embeddings (\(t\)-SNE)~\cite{van_der_maaten_visualizing_2008,van_der_maaten_learning_2009,min2010deep} to inform the training process by back-propagating the manual manipulations of the embedded latent space through the classification network.
\smallskip

% We show a way to provide a desired embedding that informs the training process for the entire representation while maintaining the overall goal of manipulating individual class performance.
% This also leads to an improved understanding of how observations in the embedded latent space relate to actual performance changes.

\textbf{Related Work:} Low dimensional representations of high dimensional latent spaces have been subject to scientific research for many decades~\cite{wold1987principal,mead1992review,tenenbaum_global_2000,van_der_maaten_visualizing_2008,mcinnes_umap:_2018}.
% The most popular approaches, in chronological order, are principal component analysis~\cite{wold1987principal}, multidimensional scaling~\cite{mead1992review}, Isomap~\cite{tenenbaum_global_2000}, \(t\)-distributed stochastic neighbourhood embedding (t-SNE)~\cite{van_der_maaten_visualizing_2008}, and Uniform Manifold Approximation~\cite{mcinnes_umap:_2018}.
Commonly these methods are treated as independent modules and applied to a selected part of the representation, e.g., the penultimate layer of a discriminator network.
However, these embeddings are often spatially inconsistent during training from epoch to epoch and cannot inform the training process through back-propagation.
Van der Maaten et al.~\cite{van_der_maaten_learning_2009,min2010deep} proposed to learn mappings through a neural network.
This approach has the advantage that it can be directly integrated into an existing network architecture enabling end-to-end forward and backward updates.
While unsupervised dimensionality reduction techniques have been used as part of deep learning workflows~\cite{tomar2014manifold,lee2015deeply,chen2017deep,rusu_meta_2018} and for visualising latent spaces~\cite{rauber_visualizing_2017,erhan_why_2010}, we are not aware of any previous work that exploited parametric embeddings for a direct manipulation of learned representations.
This shaping of the latent space relates our approach to metric learning~\cite{kulis2012metric,bellet2013survey}.
Metric learning makes use of specific loss functions to automatically constrain the latent space, but does not allow manual interventions.
\shortmethod{} are general enough to be combined with concepts of metric learning.
\smallskip

\textbf{Contribution:} We introduce \method{} (\shortmethod), a technique for (a)~understanding the relationship between a classifier and its learned latent representation, and (b) facilitating targeted performance gains by improving latent space clustering.
We discuss an application of \shortmethod{} in the context of anatomical standard plane classification during fetal ultrasound imaging.

\section{Method}

\method{} (\shortmethod{}) can be applied to any neural network classifier.
Consider a dataset \(X = \{x_1, \dots, x_N\}\) with \(N\) instances belonging to \(K\) classes.
A~neural network \(C\) was trained to predict the ground truth labels \(g_i\) of \(x_i\), where \(g_i \in \{\gamma_1, \dots, \gamma_K\}\).
Let \(C_l(x_i)\) be the activations of the network's \(l\)th layer, and let the network have \(L\) layers in total.

Given \(C\), \shortmethod{} consist of three steps:
(1)~training of a secondary network \(\tilde{E}\) that approximates a given non-linear embedding \(E = \{y_1, \dots, y_N\}\) for the outputs \(C_l(x_i)\) of layer \(l\);
(2)~modifying the positions \(y_i\) of embedded points, yielding new positions \(y^\prime_i\); and
(3)~retraining \(C\), such that \(\tilde{E}(C_l(x_i)) \approx y^\prime_i\).
In the following sections, we will discuss these three steps in detail.

\subsection{Parametric Embeddings}

% Non-linear embedding techniques based on neighbourhood graphs are routinely applied for creating low-dimensional representations of high-dimensional latent spaces~\cite{erhan_why_2010,rauber_visualizing_2017}.
% The two most widely used non-linear embedding techniques in this context is \(t\)-SNE~\cite{van_der_maaten_visualizing_2008}.

The embeddings used for \shortmethod{} are parametric approximations of \(t\)-SNE.
For \(t\)-SNE, distances between high-dimensional points \(z_i\) and \(z_j\) are converted to probabilities of neighbourhood \(p_{ij}\) via Gaussian kernels.
%rephrase in two sentences: 
The variance of each kernel is adjusted such that the perplexity of each distribution matches a given value.
This perplexity value is a smooth measure for how many nearest neighbours are covered by the high-dimensional distributions.
Then, a set of low-dimensional points is initialised and likewise converted to probabilities \(q_{ij}\), this time via heavy-tailed \(t\)-distributions.
The low-dimensional positions are then adjusted by minimising the Kullback--Leibler divergence \(\mathrm{KL}(p_{ij} || q_{ij})\) between the two probability distributions.

% \begin{equation}
%     2^{H(p_i)} = \mathrm{Perp}, \text{~where~} H(p_i) = -\sum_j p_{ij} \log_2 p_{ij}.
%     \label{eq:perp}
% \end{equation}
% The perplexity \(\mathrm{Perp}\) is a hyperparameter that can be understood as a smooth measure for how many nearest neighbors are covered by the high-dimensional distributions.

% To prevent all the points from crowding together, a heavy-tailed \(t\)-distribution is used in the embedding space instead of a normal distribution~\cite{van_der_maaten_learning_2009}.

Given a set of \(d\)-dimensional points \(z_i \in \mathbb{R}^d\), \(t\)-SNE yields a set of \(d^\prime\)-dimensional points \(z^\prime \in \mathbb{R}^{d^\prime}\).
However, it does not yield a general function \(E: \mathbb{R}^d \rightarrow \mathbb{R}^{d^\prime}\) defined for all \(z \in \mathbb{R}^d\).
It is thus impossible to add new points to existing \(t\)-SNEs or to back-propagate gradients through the embeddings.
% This lack of out-of-sample extensibility further prevents implementations that allow back-propagation of losses through the embedding.

In order to allow out-of-sample extension, van der Maaten introduced the idea of approximating \(t\)-SNE with neural networks~\cite{van_der_maaten_learning_2009}.
We adapt van der Maaten's approach and introduce two important extensions, based on recent advancements related to \(t\)-SNE~\cite{policar_opentsne:_2019}: (1)~\emph{PCA initialisation} to improve reproducibility across multiple runs and preserve global structure; and (2)~\emph{approximate nearest neighbours}~\cite{dong_2011_efficient} for a more efficient calculation of the distance matrix without noticeable effects on the embedding quality.
% \begin{itemize}
%     \item \emph{PCA initialisation:} Initialising the low-dimensional points with the first two principal components of the high-dimensional dataset leads to an improved reproducibility across multiple runs, as well as a better preservation of global structure.
%     \item \emph{Approximate nearest neighbours:} The computational overhead of calculating a full pairwise distance matrix can be circumvented without noticeable effects on the embedding quality by calculating only the distances to a given number of approximate nearest neighbours~\cite{dong_2011_efficient}.
% \end{itemize}

Our approach is an unsupervised learning workflow resulting in a neural network that approximates \(t\)-SNE for a set of input vectors \(\{z_1, \dots, z_N\}\) given a perplexity value \(\mathrm{Perp}\).
We only take into account the \(k\) approximate nearest neighbours, where \(k = \min(3 \times \mathrm{Perp}, N-1)\).
In contrast to the simple binary search used by van der Maaten~\cite{van_der_maaten_learning_2009}, we use Brent's method~\cite{brent_2013_algorithms} for finding correct variances of the kernels.
Optionally, we pre-train the network such that its 2D output matches the first two principal components of  \(z_i\).
In the actual training phase, we calculate low-dimensional pairwise probabilities \(q_{ij}\) for each input batch, and use the KL-divergence \(\mathrm{KL}(p_{ij} || q_{ij})\) as a loss function.

While van der Maaten used a network architecture with three hidden layers of sizes 500, 500, and 2000~\cite{van_der_maaten_learning_2009}, we found that much smaller networks (e.g., two hidden layers of sizes 300 and 100) are more efficient and yield more reliable results.
The \(t\)-SNE-approximating network can be connected to any complex neural network, such as CNNs for medical image classification.

\subsection{Projective Latent Constraints}

Once the network \(\tilde{E}\) has been trained to approximate the \(t\)-SNE, new constraints on the embedded latent space can be defined.
This is most easily done by visualising the embedded points, \(y_i = E(C_l(x_i))\), in a scatter plot with points coloured categorically by their ground truth labels \(g_i\).
For our applications, we chose only simple modifications of the embedding space: shifting of entire class clusters\footnote{The class cluster for class \(\gamma_j\) is simply the set of points \(\{y_i = E(C_l(x_i)) \mid g_i = \gamma_j\}\).}, and contraction of class clusters towards their centres of mass.
The modified embedding positions \(y^\prime_i\) are used as target values for the subsequent regression learning task.

In this work, we focus on class-level interventions because their effect can be directly measured via class-level performance metrics and they do not require domain-specific interactive tools that would lead to additional cognitive load.
In principle, arbitrary alterations of the embedded latent space are possible within our technique.

% For shifting the cluster corresponding to class \(\gamma_j\), we can define
% \begin{equation}
%     y^\prime_i = \left\{\begin{array}{l@{\quad}l}
%         y_i + \delta, & \text{if } g_i = \gamma_j\\
%         y_i             & \text{else}
%         \end{array}\right..
%         \label{eq:shift}
% \end{equation}
% For contraction of a class cluster for class \(\gamma_j\) by a factor of \(\kappa\) towards its centre of mass, we can define
% \begin{equation}
%     y^\prime_i = \left\{\begin{array}{l@{\quad}l}
%         (1-\kappa)\ y_i + \kappa\ \bar{y}_j, & \text{if } g_i = \gamma_j\\
%         y_i             & \text{else}
%         \end{array}\right.\text{,~where~}
%         \bar{y}_j = \mean(\{y_i \,|\, g_i = \gamma_j \}).
%         \label{eq:contract}
% \end{equation}

\subsection{Retraining the Classifier}

In the final step, the original classifier is retrained with an adapted loss function \(\mathcal{L}_\mathrm{\shortmethod{}}\) based on the modified embedding:
\begin{equation}
    \mathcal{L}_\mathrm{\shortmethod{}}(x_i, g_i, y'_i) = (1 - \lambda) \ \mathcal{L}_\mathrm{class}(C_L(x_i), g_i) + \lambda\  \mathcal{L}_\mathrm{emb}(\tilde{E}(C_l(x_i)), y^\prime_i).
    \label{eq:loss}
\end{equation}
The new loss function combines the original classification loss function \(\mathcal{L}_\mathrm{class}\), typically a cross-entropy term, with an additional term \(\mathcal{L}_\mathrm{emb}\).
Minimisation of \(\mathcal{L}_\mathrm{emb}\) causes the classifier to learn new activations that yield embedded points similar to \(y^\prime\) (using the given embedding function \(\tilde{E}\)).
As \(\tilde{E}\) is simply a neural network, back-propagation of the loss is straightforward.
In our experiments, we use the squared euclidean distance for \(\mathcal{L}_\mathrm{emb}\) and test different values for the weighting coefficient \(\lambda\).
We also experiment with only counting the embedding loss for instances of classes that were altered in the embedding.

\section{Experiments}

\subsection{MNIST and CIFAR}

\begin{figure}[t]
    \centering
    \includegraphics[width=.95\textwidth]{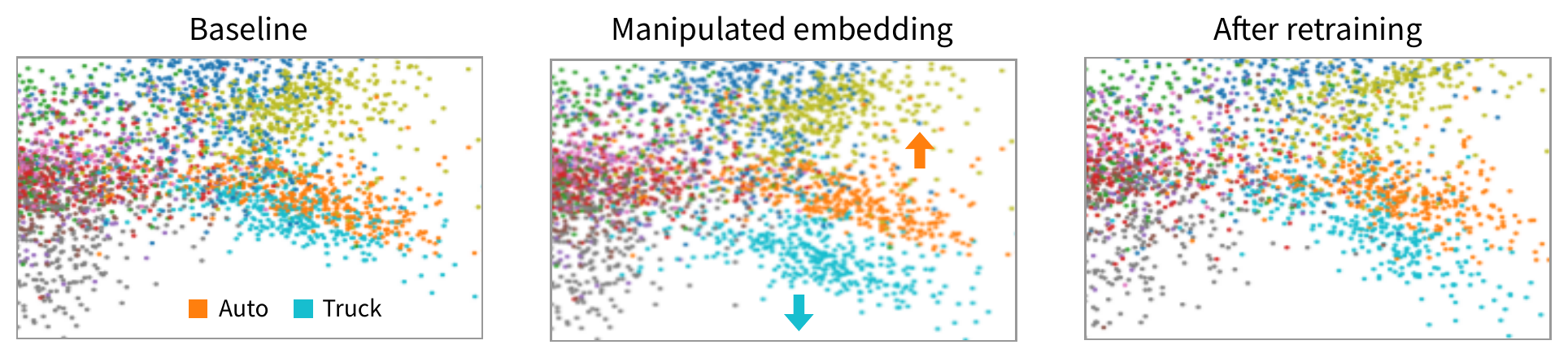}
    \caption{Detail views of the embedded latent space before (left), during (centre) and after (right) \method{} for classification of CIFAR-10 images, focusing on the classes \emph{Truck} and \emph{Auto}.}
    \label{fig:cifar-plot}
\end{figure}

As a proof of concept, we applied \shortmethod{} to simple image classifiers: a small MLP for MNIST~\cite{lecun_mnist_2005} images and a simple CNN for CIFAR-10~\cite{CIFAR-10} images.
For MNIST, the embedded latent space after retraining generally preserved the manipulations well, when class clusters were contracted and/or translated.
The classification accuracy only changed insignificantly (within a few percent over wide ranges of \(\lambda\)).
Typical results for the CIFAR-10 classifier are shown in Fig.~\ref{fig:cifar-plot}, where the goal of the \method{} was to reduce the model's confusion between the classes \emph{Truck} and \emph{Auto}, by separating the respective class clusters.
When comparing a classifier trained for \(5 + 4\) epochs with \(\mathcal{L}_\mathrm{class}\) to one trained for 5 epochs \(\mathcal{L}_\mathrm{class}\) \(+\) 4 epochs \(\mathcal{L}_\mathrm{\shortmethod}\), the latter showed a relative increase of target-class-specific \(F_1\)-scores by around 5\,\%, with the overall accuracy improving or staying the same.
The embeddings after retraining, as seen in Fig.~\ref{fig:cifar-plot}, reflected the manual interventions well, but not as closely as in the case of MNIST.
We also found that, in the case of CNNs, using the activations of the final dense layer (\(l=L\)) yielded the best results.

\subsection{Standard Plane Detection in Ultrasound Images}

We tested our approach on a challenging diagnostic view plane classification task in fetal ultrasound screening. 
The dataset consists of about 12,000 2D fetal ultrasound images sampled from 2,694 patient examinations with gestational ages between 18 and 22 weeks.
Eight different ultrasound systems of identical make and model (GE Voluson E8) were used for the acquisitions to eliminate as many unknown image acquisition parameters as possible.
Anatomical standard plane image frames were labelled by expert sonographers as defined in the UK FASP handbook~\cite{screening2015}.
We selected a subset of images that tend to be confused by established models~\cite{baumgartner2017sononet}: Four Chamber View (4CH), Abdominal, Femur, Spine, Left Ventricular Outflow Tract (LVOT) and Right Ventricular Outflow Tract (RVOT)\,/\,Three Vessel View (3VV).
RVOT and 3VV were combined into a single class after clinical radiologists confirmed that they are identical.
We split the resulting dataset into 4,777 training and 1,024 test images.

\begin{figure}
    \centering
    \includegraphics[width=.99\textwidth]{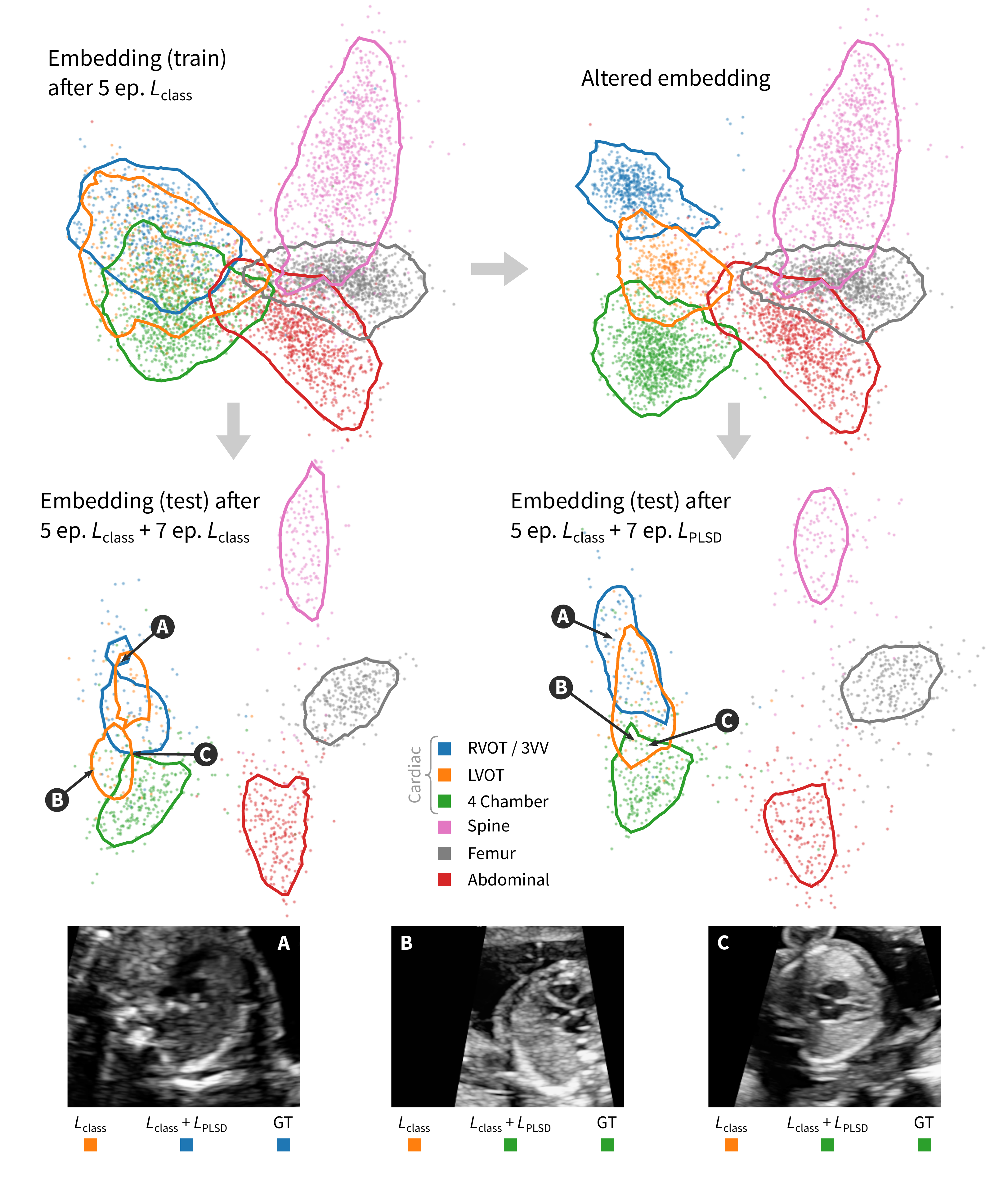}
    \caption{\method{} for standard plane classification in fetal ultrasound images.
    Top left: embedding of the baseline network's output (train) after 5~epochs of classification training (\(\mathcal{L} = \mathcal{L}_\mathrm{class}\)).
    Top right: altered output embedding (train) with manually separated cardiac classes.
    Centre left: Output embedding (test) after resuming standard classification training for 7~epochs (\(\mathcal{L} = \mathcal{L}_\mathrm{class}\)), starting from the baseline classifier (top left).
    Centre right: embedding (test) after resuming training with an updated loss function (\(\mathcal{L} = \mathcal{L}_\mathrm{\shortmethod{}}=0.9\ \mathcal{L}_\mathrm{class} + 0.1\ \mathcal{L}_\mathrm{emb}\)), starting again from the baseline classifier (top left).
    For easier comparability, class-specific contour lines at a density threshold of \(1/N\) are shown, where \(N\) is the total number of train or test images, respectively.
    Performance measures for the classifiers are given in Table~\ref{tab:sononet}.
    Bottom: Three example images that were successfully classified after applying PLSD.
    For each image, the positions in both embeddings are indicated.
    }
    \label{fig:sononet-plot}
\end{figure}

The architecture of our baseline classifier is SonoNet-64~\cite{baumgartner2017sononet}.
The network was trained for 5 epochs with pure classification loss, i.e., \(\mathcal{L} = \mathcal{L}_\mathrm{class}\).
We used Kaiming initialization, a batch size of 100, a learning rate of 0.1, and 0.9 Nesterov momentum.
During these first five training epochs, we used random affine transformations for data augmentation (\(\pm15^\circ\)~rotation, \(\pm0.1\)~shift, 0.7 to 1.3~zoom).

The 6-dimensional final-layer logits for the non-transformed training images were used as inputs for the training of the parametric \(t\)-SNE network.
We used a fully connected network with two hidden layers of sizes 300 and 100.
The \(t\)-SNE network was trained for 10 epochs with a learning rate of 0.01, a batch size of 500 and a perplexity of 50.
We pre-trained the network for 5 epochs to approximate a PCA initialisation.

The ultrasound dataset is imbalanced, with 1,866 images in the three cardiac classes, and 2,911 images in the three non-cardiac classes.
There are about twice as many 4CH images as RVOT/3VV, and three times as many 4CH images as LVOT.
As a result, after five epochs of classification learning, our vanilla classifier could not properly distinguish between the three cardiac classes.
This is apparent in the baseline embedding shown in Fig.~\ref{fig:sononet-plot} (top left).

We experimented with \shortmethod{} to improve the performance for the cardiac classes, in particular for RVOT/3VV and LVOT.
Figure~\ref{fig:sononet-plot} (top right) shows the case of contracting and shifting the class clusters of RVOT/3VV and LVOT.
% by a factor of \(\kappa = 1/2\), and that of 4CH by a factor of \(\kappa = 1/4\), according to Eq.~\ref{eq:contract}.
% Additionally, the three clusters were shifted by \(\delta_\mathrm{RVOT} = (-7,+10)\), \(\delta_\mathrm{4CH} = (-5,-10)\), and \(\delta_\mathrm{LVOT} = (+2,-2)\), respectively, according to Eq.~\ref{eq:shift}.

\begin{table}[t]
    \newcommand{\z}{\bfseries}
    \centering
    \caption{Global and class-specific performance measures for standard plane classification in fetal ultrasound images with and without \shortmethod{}, evaluated on the test set. The last two columns are weighted averages of the values for the three cardiac and the three non-cardiac classes, respectively. (*~The class labelled as RVOT also includes 3VV.)}
    \footnotesize
    \begin{tabular}{l@{\quad}l@{\quad}*{5}{c@{~~}}c@{\quad}c@{~~}c}
        \toprule
        & & RVOT* & 4CH & LVOT & Abd. & Femur & Spine & Cardiac & Other \\
        \cmidrule(r){3-8}\cmidrule(l){9-10}
        Precision
        & Class. only &
        \z0.82 & 0.82 & 0.42 & 0.93 & 0.98 & 0.97 & 0.77 & 0.96 \\
        & \shortmethod{} &
        0.78 & \z0.85 & \z0.61 & 0.91 & 0.97 & 0.96 & \z0.80 & 0.95 \\
        \cmidrule(r){3-8}\cmidrule(l){9-10}
        Recall
        & Class. only &
        0.38 & 0.94 & \z0.46 & 0.96 & 0.97 & 0.94 & 0.76 & 0.96 \\
        & \shortmethod{} &
        \z0.73 & 0.94 & 0.28 & 0.96 & 0.97 & 0.94 & \z0.81 & 0.96 \\
        \cmidrule(r){3-8}\cmidrule(l){9-10}
        \(F_1\)-score
        & Class. only &
        0.56 & 0.88 & \z0.44 & 0.95 & 0.97 & 0.95 & 0.75 & 0.96 \\
        & \shortmethod{} &
        \z0.76 & \z0.89 & 0.41 & 0.94 & 0.97 & 0.95 & \z0.80 & 0.95 \\
        \bottomrule
    \end{tabular}
    \label{tab:sononet}
\end{table}
    
After the latent interventions, training was resumed for 7 epochs with the mixed loss function defined in Eq.~\ref{eq:loss}.
We experimented with different values for \(\lambda\); all results given in this section are for \(\lambda = 0.1\), which was found to be a suitable value in this application scenario.
For a fair comparison, training of the baseline network was also resumed for 7 epochs with pure classification loss.
In both cases, the remaining training epochs were performed without data augmentation, but with all other hyperparameters kept the same as for the vanilla classifier.

The outputs were then embedded with the parametric \(t\)-SNE learned on the baseline outputs (see Fig.~\ref{fig:sononet-plot}, centre).
By resuming the training with included embedding loss, the clusters for the three cardiac classes assume relative positions that are closer to those in the altered embedding.
The contraction constraint also led to more convex clusters for the test outputs.
Figure~\ref{fig:sononet-plot} (bottom) shows three exemplary images that were misclassified in case of the pure classification loss model, but correctly classified after applying \shortmethod{}.
Further inspection showed that most of the images that were correctly classified after \shortmethod{} (but not before) had originally been embedded close to decision boundaries.

Table~\ref{tab:sononet} lists the class-specific precision, recall, and \(F_1\)-scores for the two different networks.
By applying \shortmethod{}, the average quality for the cardiac classes could be improved without negatively affecting the performance for the remaining classes.
In some experiments, we observed much larger quality improvements for individual classes.
For example, in one case the \(F_1\)-score for LVOT improved by a factor of two.
In these extreme cases, however, local improvements were often accompanied by significant performance drops for other classes.
%We also tried to declutter RVOT and 3VV in case of a specialized classifier for only the cardiac classes, but found that \shortmethod{} was not able to improve these classes significantly.

\section{Discussion}
\vspace*{-.25ex}

The insights gained from \shortmethod{} about the relationship between a classifier and its latent space are based on an assessment of the model's response to the interventions.
This response can be evaluated on two axes: the \emph{embedding response} and the \emph{performance response}.

Simple classifications tasks, for which the baseline classifier already works well (e.g., MNIST) often show a considerable embedding response with only a minor performance response.
This means that the desired alterations of the latent space are well reflected after retraining without strong effects on the classification performance.
Such classifiers are flexible enough to accommodate the latent manipulations, likely because they are overparameterised.
In more complex cases, such as CIFAR, the embedding response is weaker, but often accompanied by a more pronounced class-specific performance increase.
For these cases, the learned representation seems to be more rigidly connected with the classification performance.
Finally, the standard plane detection experiments showed that sometimes a minimal change in the embedding is accompanied by a considerable performance increase for the targeted classes.
Here, the overall structure of the embedding seems to be fixed, but the classification accuracy can be redistributed between classes by injecting additional domain knowledge while allowing non-targeted classes to move freely.

In general, we found that too severe alterations of the latent space cannot be preserved well since the embeddings are based on local information.
Furthermore, seemingly obvious changes made in the embedding may contradict the original classification task due to the non-linearity of the embedding.
The strength of \shortmethod{} is that a co-evaluation of the two components of the loss function can reveal these discrepancies.
As a result, even when \shortmethod{} cannot be used for improving a classifier's performance, it can still lead to a better understanding of the flexibility of the model and/or the trustworthiness of the embedding.

% Second, in order to learn the alterations correctly for some classes, other classes need to be allowed to move more freely.
% In the ultrasound experiments, for example, we found that restricting the embedding loss only to the cardiac classes---while leaving the remaining classes free to move---led to the best results.
% Finally, the embedding itself is non-linear and, despite the PCA initialisation, might only represent a local minimum of the Kullback--Leibler divergence.
% This means that seemingly obvious changes made in the embedding often completely contradict the original classification task.
% Thus, both components of the \shortmethod{} loss need to be observed carefully during training.

%These guidelines are particularly important in challenging real-world scenarios.
% In case of easier classification tasks, such as MNIST~\cite{lecun_mnist_2005} and CIFAR-10~\cite{CIFAR-10}, we found that the process was much more forgiving towards more extreme changes of the embedding.
% We also found that relatively high values for \(\lambda\) still led to improved classification performance.

In future work, we would like to experiment with parametric versions of different dimensionality reduction techniques and explore the potential of instance-level manipulations controlled via an interactive visualisation.

\enlargethispage{\baselineskip}

\section{Conclusion}

We introduced \method{}, a promising technique to inject additional information into neural network classifiers by means of constraints derived from manual interventions in the embedded latent space.
\shortmethod{} can help to get a better understanding of the relationship between the latent space and a classifier's performance.
We applied \shortmethod{} successfully to obtain a targeted improvement in standard plane classification for ultrasound images without negatively affecting the overall performance.

%While PCA would be a trivial candidate, as it directly yields a linear mapping, it typically suffers from poor clustering performance.
%More interesting for future experiments are parametrized version of Isomap and UMAP, on which we are currently working.

% The source code of \shortmethod{} will be publicly available by the time of the conference.

%\shortmethod{} facilitates finding the most suitable embedding manipulations, and is suitable to be directly included in whole ML workflows.

\bigskip
\noindent
\begingroup\small
\textbf{Acknowledgments:} 
This work was supported by the State of Upper Austria (Human-Interpretable Machine Learning) and the Austrian Federal Ministry of Education, Science and Research via the Linz Institute of Technology (LIT-2019-7-SEE-117), and by the Wellcome Trust (IEH 102431 and EPSRC EP/S013687/1.).
\endgroup

% {\small This work was supported by the State of Upper Austria (Human-Interpretable ML), by the Linz Institute of Technology (LIT-2019-7-SEE-117), and by the Wellcome Trust (IEH 102431 and EPSRC EP/S013687/1.)}

\newpage

\bibliographystyle{splncs04}
\bibliography{references}

\end{document}